\pdfoutput=1

\documentclass[11pt]{article}

\usepackage{acl}
\usepackage{times}
\usepackage{latexsym}
\usepackage{caption}
\usepackage{subcaption}
\usepackage{ulem}
\usepackage{booktabs}
\usepackage[T1]{fontenc}

\usepackage[utf8]{inputenc}

\usepackage{microtype}

\usepackage{inconsolata}
\usepackage{adjustbox}

\title{Argument Mining in Data Scarce Settings: Cross-lingual Transfer and Few-shot Techniques}

 \author{Anar Yeginbergen \and Maite Oronoz \and Rodrigo Agerri \\
         HiTZ Center - Ixa, University of the Basque Country UPV/EHU\\
         \texttt{\{anar.yeginbergen,maite.oronoz,rodrigo.agerri\}@ehu.eus}
         }

\begin{document}
\maketitle
\begin{abstract}

Recent research on sequence labelling has been exploring different strategies to mitigate the
lack of manually annotated data for the large majority of the world languages. Among others, the
most successful approaches have been based on (i) the cross-lingual transfer capabilities of multilingual
pre-trained language models (model-transfer), (ii) data translation and label projection (data-transfer) and (iii), prompt-based learning by reusing the mask objective to exploit the few-shot capabilities of pre-trained language models (few-shot). Previous work seems to conclude that model-transfer outperforms data-transfer methods and that few-shot techniques based on prompting are superior to updating the model's weights via fine-tuning.  In this paper, we empirically demonstrate that, for Argument Mining, a sequence labelling task which requires the detection of long and complex discourse structures, previous insights on cross-lingual transfer or few-shot learning do not apply. Contrary to previous work, we show that for Argument Mining data transfer obtains better results than model-transfer and that fine-tuning outperforms few-shot methods. Regarding the former, the domain of the dataset used for data-transfer seems to be a deciding factor, while, for few-shot, the type of task (length and complexity of the sequence spans) and sampling method prove to be crucial.
\end{abstract}

\section{Introduction}\label{sec:introduction}
Transfer learning and pre-trained language models are closely related as the knowledge learned for one or more tasks in one specific language can be applied to other tasks or languages \cite{WANG:2023}. In this paper, we analyze how this feature can be applied in scenarios where not much data is accessible as it is the case of argument mining in the clinical domain. In \textit{data-transfer} approaches, data can be translated and the required annotations projected to train supervised models. \textit{Model-transfer} methods avoid the long process of generating the training data by applying multilingual pre-trained language models to learn the annotations in one language and generate the predictions in a different one \cite{PIKULIAK:2021,garcia2022model,chen-etal-2023-frustratingly}. Alternatively, by \textit{few-shot} prompting there is a possibility to reach comparable results by providing a few examples from the problem at hand to pre-trained language models \cite{ma-etal-2022-template}. In sequence labelling tasks, these methods have shown to be effective with a minimal loss in performance based on a very few annotated examples.

These \textit{few-shot} methods have widely been tested on popular benchmark datasets, such as in those for Named Entity Recognition (NER) (CoNLL 2003 \cite{tjong-kim-sang-de-meulder-2003-introduction}, OntoNotes 5.0 \cite{AB2/MKJJ2R_2013}, MIT-Movie \cite{6707708}) concluding that model-transfer outperform data-transfer methods and that few-shot techniques based on prompting are superior to updating the model's weights via fine-tuning. However, such conclusions have been based on results obtained on sequence labelling tasks for which the sequence spans are commonly short and quite homogeneous in terms of the structure and content of the label words.

In this paper we explore whether these conclusions still hold for Argument Mining, a task in Natural Language Processing (NLP) aimed at extracting long and complex discourse structures from text.  Argument Mining usually involves two distinct subproblems: (1) \textit{argument component detection}, focusing on locating the spans of arguments and identifying their types (e.g., claims and premises), and (2) \textit{classification of argument relations}, which involves classifying the relationship between two argument components as \textit{supporting} or \textit{attacking}.

In order to do so, we use AbstRCT \citep{mayer2021enhancing} a corpus of medical abstracts annotated for the detection of argument components. The original corpus is published in English and has been extended it into a parallel multilingual corpus of medical arguments in Spanish, Italian, and French\footnote{\url{https://huggingface.co/datasets/HiTZ/multilingual-abstrct}} by translating with state-of-the-art language models and projecting the annotations to the target languages using the technique of \citet{garcia2022model}. 

Summarizing, we investigate the following two research questions to address data scarcity in Argument Mining:

\begin{itemize}
    \item \textbf{RS1}: What approach is better to overcome data scarcity: data-transfer, model-transfer or few-shot learning?
    \item \textbf{RS2}: What is the influence of the type of task (length and complexity of the sequence spans) and sampling methods for optimal results in few-shot settings?  
\end{itemize}

In this paper we empirically demonstrate that, for Argument Mining (AM), a sequence labelling task that requires the detection of long and complex discourse structures, previous insights on cross-lingual transfer or few-shot learning do not apply. Contrary to previous work, we show that for Argument Mining data-transfer obtains better results than model-transfer and that fine-tuning outperforms few-shot methods. Regarding the former, the domain of the dataset used for data-transfer seems to be a deciding factor, while, for few-shot, the type of task (length and complexity of the sequence spans) and the sampling method proves to be crucial. Data and code for the experiments described in this paper are publicly available in: \url{https://github.com/anaryegen/few_shot_argument_mining}.

\section{Related Work}\label{sec:related}
In this section, we review the closest work to the paper's main topics, namely, Argument Mining, cross-lingual transfer and few-shot learning. 

\subsection{Argument Mining}

The are a number of different theoretical approaches to describe the argument structures that can be inferred from text analysis. For instance, \citet{toulmin1958uses} identified different functional roles in arguments (evidence, warrant, backing, qualifier, rebuttal, and claim) based on how the conclusion is drawn from evidence in the text. Furthermore, \citet{freeman2011argument} investigated how to transfer arguments via diagramming techniques of the informal logic tradition. Others \cite{dung1995acceptability} tried to create a graph-based representation of argumentation by applying non-monotonic reasoning in Artificial Intelligence (AI) and logic programming. Finally, \citet{peldszus2013argument} introduced a diagram structure with models of the textual representation of arguments and globally optimized argumentative relations. They argued that \textit{support} and \textit{attack} relations are sufficient to describe the overall relationships between argument components. Moreover, they identified five different types of argument graphs based on the connections that exist between them, namely, one claim having relations with multiple premises, a claim followed by another claim, etc.

In Natural Language Processing Argument Mining (AM) is focused on automatically identifying the argument components and classifying the relations that may exist between them. Following the theoretical models proposed, a number of empirical approaches have been developed in the last few years. Thus, \citet{stab-gurevych-2017-parsing} tackled AM in two different steps. First, they try to locate the span argumentative text and classify the type of component at token level. Second, they classify the relations linking the identified argument spans. In addition to the two step system to address AM, they also generate Persuasive Essays, perhaps the most popular NLP dataset manually annotated with argument structures \cite{stab-gurevych-2017-parsing}. Later on, \citet{eger-etal-2017-neural} introduced an end-to-end AM system based on a bi-directional sequence-to-sequence model. 

Other work includes \citet{toledo-ronen-etal-2020-multilingual}, which provides an in detail analysis at argument level of various multilingual datasets, while \citet{rocha2018cross} experimented with cross-lingual argumentative relation identification from English to Portuguese.

Finally, \citet{mayer2020ecai} introduced the first dataset of English medical abstracts annotated for argument component detection and argument relation classification. Subsequently, \citet{mayer2021enhancing} introduced a Transformer-based solution with Gated Recurrent Units (GRU) and Conditional Random Field (CRF) classification layers.

\subsection{Few-shot Learning Approaches for Sequence Labelling}

The availability of pre-trained language models allows to apply supervised methods with less amount of annotated data which is why some research in different NLP tasks has focused on few-shot training \citep{hofer2018few, 10.1145/3297280.3297378, 9262018}, namely, learning supervised models with very few manually annotated samples. The rise of prompt-based models \citep{radford2019language, brown2020language} further increased the interest in learning the task describing the classification objective. This usually involves transforming traditional classification tasks into cloze tasks using textual templates and a predefined set of label words, highlighting the importance of template design in prompt-based learning. 

In this line of work, \citet{schick-schutze-2021-exploiting}  presented a semi-supervised training approach that reformulates input instances into cloze-style phrases. \citet{cui-etal-2021-template} proposed a template-based method for Named Entity Recognition (NER) by generating templates for each entity from a given example. However, template-based approaches are better suited to sentence-level tasks where the complexity of the templates remain manageable. As an alternative, EntLM \cite{ma-etal-2022-template} proposed a template-free few-shot learning approach for sequence labelling tasks. Their method is based on computing a set of label words from the input text and replacing the entity-specific tokens with these label words in the training sample. EntLM obtains state-of-the-art results which is why we use it in this paper as the representative of few-shot learning for argument component detection. \citet{huang-etal-2022-copner} and \citet{das-etal-2022-container} propose few-shot learning for NER involving contrastive learning via prompt-based meta-learning. However, their methods require large amounts of data to first train the model before adapting it with a handful of examples for various label sets.

\subsection{Cross-lingual Sequence Labelling}

Previous work on cross-lingual sequence tagging mainly focuses on tasks such as part-of-speech (POS) tagging, named-entity-recognition (NER) \citep{gaddy2016ten, yang2017transfer, agerri2018building, chen2018multi, liu2020importance}, and Opinion Target Extraction (OTE) \citep{agerri2019language}.
\citet{garcia2022model} compared model-transfer and data-transfer approaches on a variety of sequence labelling tasks, datasets, and languages. They conclude that model-transfer using pre-trained multilingual language models such as XLM-RoBERTa-large \cite{conneau2019unsupervised} outperform data-transfer methods.

Closer to our work, \citet{eger-etal-2018-cross} generated parallel German and Chinese versions from English by applying manual and automatic translation and label projection to experiment with \textit{data-transfer} approaches based on cross-lingual embeddings. They concluded that, while machine translated data degraded results when used for training a supervised model for the target language, results were promising enough to continue working on that research direction. Thus, \citet{sousa2021cross} translated Persuasive Essays into Portuguese for further cross-lingual experimentation. However, it should be noted that current \textit{model-transfer}, \textit{few-shot} and supervised techniques based on multilingual pre-trained language models are clearly superior to the methods used at the time, which makes the purpose of our work rather relevant.

\section{Data}\label{sec:data}

The starting point for experimentation on argument mining in data scarce settings is AbstRCT, a dataset of Randomized Controlled Trials (RCT) manually annotated with argument components and relations \cite{mayer2021enhancing}. The original AbstRCT consists of abstracts of clinical trials in English collected from the MEDLINE database and manually annotated with two types of argument components: \textit{Claims} and \textit{Premises}. A `claim'  is a concluding statement about the outcome of the study. In the medical domain it typically refers to a judgement regarding a possible diagnosis or a treatment. A `premise' corresponds to an observation or measurement in the study (ground truth), which supports or attacks another argument component, usually a claim. It is important to stress that premises are observed facts, therefore, credible without further evidence.

The training set consists of 350 abstracts that cover the neoplasm disease, 50 more abstracts about neoplasm are used for development, while the three evaluation sets are composed of: 100 abstracts about neoplasm, 100 abstracts about glaucoma and finally a mixed set of 100 abstracts with 20 abstracts for each of the diseases in the AbsRCT dataset (i.e. neoplasm, glaucoma, hypertension, hepatitis and diabetes). The number of the sequences with Premise and Claim argument components in these sets is shown in Table \ref{table:arg_comp_dist}.

\begin{table}[ht]
\begin{center}
\begin{tabular}{ lccc } 
 \hline
Data & \# of Premise & \# of Claim \\ 
 \hline
 Train: Neoplasm    & 1535 & 730 \\
 Dev: Neoplasm      & 438  & 228 \\
 Test: Neoplasm & 438  & 248 \\
 Test: Glaucoma & 404  & 190 \\
 Test: Mixed    & 388  & 212  \\
 \hline
\end{tabular}
\caption{Number of sequences with Premise and Claim argument components in the train, dev, and test sets.}
\label{table:arg_comp_dist}
\end{center}
\end{table}

We machine-translated with the state-of-the-art machine translation model No Language Left Behind (NLLB) \cite{costa-etal-2022-domain} into Spanish, Italian, and French. Subsequently, we projected the annotations from the original dataset into the translated versions using the annotation projection tool developed by \citet{garcia2022model}. In the last phase, native speakers manually corrected the projections of the argument component labels. This was required to have gold standard evaluation data. While it would had been interesting to project the dataset to other languages, we only had in-house expertise to manually check the annotations for Spanish, Italian and French.

We also generated a \textit{post-processed} version by programatically correcting systematic errors performed during the automatic projection of the annotations. This \textit{post-processed} version fixed relatively simple but repetitive issues such as omitting the labelling of articles as argument types. As a result, we obtained three versions of the projected data: \textit{auto projected}, \textit{post-processed} and \textit{manually corrected}.

Table \ref{table:projection_f1} reports the evaluation of the \textit{auto-projected} and \textit{post-processed} annotations with respect to the gold standard (manually corrected). Results show that \textit{manually corrected} data is crucial at least for evaluation although the \textit{post-processed} version of the projections gets close enough to the gold standard.

\begin{table}
\begin{center}
\resizebox{7.5cm}{!}{
\begin{tabular}{cccc} 
  \textbf{Test set}  & \textbf{Spanish} & \textbf{French} & \textbf{Italian}  \\
  \hline
   \textit{auto-projected} &&& \\
 \hline
 Neoplasm  & 83.95 & 94.18 & 92.44   \\
 Glaucoma  & 67.97 & 90.43 & 93.79   \\
 Mixed     & 83.45 & 90.89 & 91.42   \\
   \hline
   \textit{post-processed} &&& \\
 \hline
 Neoplasm  & 95.54 & 97.87 & 98.97   \\
 Glaucoma  & 97.88 & 97.89 & 99.41  \\
 Mixed     & 95.78 & 96.97 & 97.65   \\
\hline
\end{tabular}
}
\caption{F1-score of auto-projected and post-processed data compared with manually corrected data in Spanish, French, and Italian.} 

\label{table:projection_f1}
\end{center}
\end{table}

The full training data is used for multilingual and cross-lingual experiments. To perform few-shot experiments the data is randomly sampled following different sampling approaches.

\subsection{Sampling Data for Few-shot Learning}

The main objective of Few-Shot Learning (FSL) is to generalize while learning from a small portion of data. In order to perform FSL, the data is sampled into smaller subsets and provided to the model. While state-of-the-art methods on few-shot for sequence labelling have been focused on the training method, they have not usually paid any attention to the data sampling technique \cite{ma-etal-2022-template}. In this paper, we demonstrate the importance of 
data sampling for a sequence labelling task such as Argument Mining.

We sample the data in two ways, using a method called \textit{k-shot} (based on \citet{ma-etal-2022-template}) and another one named \textit{k-percent}, where $\textit{k} \in \{ 5, 10, 20, 50 \} $. In the \textit{k-shot} method, each of the subsets contains exactly \textit{k} argument component sequences of \textit{Claim} and \textit{Premise}.
With the \textit{k-percent} sampling method we calculate the \textit{k} proportion for each argument component from the full data to reflect the distribution. 
The distribution of the sequences sampled with \textit{k}-percent method and \textit{k}-shot are shown in Table \ref{tab: kshot_arguments_distribution}.
The sequences in every sample are selected randomly in a greedy manner.

AbstRCT contains texts annotated with labels \textit{Claim}, \textit{Premise}, and \textit{O} (Outside). One sentence could belong to one or more argument component classes from the beginning until the end. In many sequence labelling tasks, the span of the components to predict consists of several words that make up only a part of the sentence, whereas in argument mining argument components can constitute a whole sentence. Hence, for the few-shot training, it is crucial to include examples without any argument components separately, namely, examples in which every token in the sequence is labeled with the \textit{O} class. If such examples are not included, the few-shot model fails to learn to classify sequences as non-arguments. 

\begin{table}[ht]
    \begin{adjustbox}{max width=7.5cm}
    \centering
    \begin{tabular}{c|ccccc}
        \textit{K} & \textbf{B-Claim} &\textbf{B-Premise} & \textbf{I-Claim} & \textbf{I-Premise} & \textbf{O}  \\
         \hline
         5 shot & 5 & 5 & 108 & 165 & 143 \\
         10 shot & 10 & 10 & 187 & 273 & 258 \\
         20 shot & 20 & 20 & 348 & 554 & 594 \\
         50 shot & 50 & 50 & 1000 & 1371 & 1389 \\
         \hline
         5\% & 36 & 76 & 712 & 2111 & 3106 \\
         10\% & 73 & 153 & 1421 & 4231 & 6108 \\
         20\% & 146 & 307 & 2832 & 8308 & 12252 \\
         50\% & 365 & 767 & 7283 & 21205 &  30322 \\
         100\% & 730 & 1535 & 14396 & 42466 & 61173 \\
        \hline
    
    \end{tabular}
    \end{adjustbox} 
    \caption{Average number of token-level Argument Components with \textit{k-shot} and \textit{k-percent} sampling in the English training set among 3 sampled files for each \textit{k}-sample.}
    \label{tab: kshot_arguments_distribution}
\end{table}

The data has a sentence-by-sentence split, where each token in the sentence is annotated with the labels following the IOB2 schema, meaning that the beginning of the argument is tagged as \textit{B-} followed by the argument component class name (Claim or Premise), the rest of the argumentative tokens are labelled with \textit{I-}, and non-argumentative sequences are labelled as \textit{O}. Since one sentence holds one or more argument types, and they tend to be lengthy, a considerable imbalance between \textit{B-} and \textit{I-} tokens is created. In Table \ref{tab: kshot_arguments_distribution}, 
we provide the distribution of the data at token level to show the imbalance in the number of tokens that are marked as \textit{B-}, \textit{I-} or \textit{O}.

Along with sampling the training data for each language, we additionally merge all the training sets from every \textit{k}-percent sampling into one to perform multilingual experiments. Therefore, the multilingual \textit{k}-percent sample is a combination of \textit{k} examples from each language from the \textit{k}-percent sample.


\section{Experimental Setup}\label{sec:method}

An important feature of AM with respect to other sequence labelling tasks is that arguments are considerably long and composed by a variety of word types.  

The experiments are based on the three different techniques that we will be comparing to establish which one is the optimal one for AM in data-scarce settings: (i) data-transfer, (ii) model-transfer and (iii), few-shot learning for sequence labelling.


Results are reported using F1 macro-averaged score calculated at sequence level, namely, the F1-score is computed for each argument component following the usual method for sequence labelling tasks as formulated for Named Entity Recognition \cite{tjong-kim-sang-de-meulder-2003-introduction}.

\subsection{Data-Transfer and Model-Transfer}
 

\textit{Data-transfer} involves generating training data in the target language by translating and projecting the annotations from the original English language to Spanish, French and Italian. This process was described in Section \ref{sec:data}. The translated and projected training data is then used to fine-tune pre-trained encoder language models.

Initially, we separately fine-tune multilingual BERT \cite{devlin-etal-2019-bert}, on the training sets of English, Spanish, French, and Italian AbstRCT corpora and evaluate the resulting models for each of the languages in a \textit{monolingual setting}\footnote{Preliminary experimentation showed that mBERT outperformed other multilingual encoder-only models such as XLM-RoBERTa or mDeBERTa-v3-base. See mDeBERTa results in Appendix \ref{sec:appendix_mono_multi_cross_mdeberta}.}.

We also tested \textit{data-transfer} in a \textit{multilingual} setting by fine-tuning multilingual BERT on the training sets for the 4 languages. Finally, both \textit{monolingual} and \textit{multilingual} settings were evaluated using both \textit{post-processed} and \textit{manually corrected} versions of the data (French, Italian and Spanish).


\textit{Model-transfer} is facilitated by pre-trained multilingual language models such as mBERT by enabling them to label sequences in languages on which they have not been explicitly trained on, relying on their multilingual or crosslingual abilities. Thus, \textit{model-transfer} allows to perform AM for languages for which no annotated data is available by training in English and generating predictions in the target language (French, Italian and Spanish). In our experiments, this amounts to fine-tuning mBERT using English data and evaluating its performance on test data from the other three languages.

\begin{table*}
\begin{center}
\footnotesize{
\begin{tabular}{cccccc} 
  \textbf{Test set} & \textbf{English} & \textbf{Spanish} & \textbf{French} & \textbf{Italian} & \textbf{Avg.} \\
  \toprule
 & \textbf{gold} & \multicolumn{3}{c}{\textbf{monolingual data-transfer}}\\
 \midrule
 Neoplasm & 61.34(1.83) & 58.54(0.49) & 60.28(1.57) & 57.29(1.12) & 59.36 \\
 Glaucoma & 64.35(0.81) & 60.63(1.56) & 64.81(2.64) & 61.95(1.18) & 62.94 \\
 Mixed    & 60.57(2.33) & 57.27(1.36) & 57.79(1.07) & 56.84(0.51) & 58.12 \\
\midrule

 & \textbf{gold} & \multicolumn{4}{c}{\textbf{monolingual data-transfer (post)}}\\
 \midrule
 Neoplasm & 61.34(1.83) & 58.88(1.76) & 55.79(1.68) & 57.64(1.63) & 57.44 \\
 Glaucoma & 64.35(0.81) & 62.86(1.48) & 62.24(1.53) & 62.37(1.74) & 62.49 \\
 Mixed    & 60.57(2.33) & 57.92(0.72) & 55.75(2.01) & 55.54(1.77) & 56.40 \\

\midrule 
\multicolumn{6}{c}{\textbf{multilingual data-transfer}} \\
\midrule
 Neoplasm & 61.89(1.41) & 59.96(1.79) & 61.17(2.25) & 59.95(2.29) & \textbf{60.74} \\
 Glaucoma & 66.97(2.04) & 65.94(1.19) & 67.14(1.62) & 60.69(0.99) & 65.19 \\
 Mixed    & 62.28(0.81) & 60.86(1.96) & 60.68(1.67) & 60.08(2.68) & \textbf{60.98} \\

 \midrule
 \multicolumn{6}{c}{\textbf{multilingual data-transfer (post)}} \\
 \midrule
  Neoplasm & 55.86(2.16) & 58.89(2.82) & 59.19(0.97) & 58.03(1.67) & 59.50 \\
  Glaucoma & 64.86(1.31) & 66.98(2.07) & 64.65(2.35) &  66.24(1.36) & \textbf{66.21} \\
  Mixed    & 57.65(2.59) & 58.49(0.70) & 58.72(2.07) & 58.06(0.66) & 59.39 \\
  \midrule
 \multicolumn{6}{c}{\textbf{cross-lingual model-transfer}} \\
 \midrule
 Neoplasm & - & 55.80(1.04) & 53.75(1.32) & 50.83(0.60) & 55.43 \\
 Glaucoma & - & 58.39(1.57) & 57.25(1.48) & 56.52(0.77) & 59.13 \\
 Mixed    & - & 52.25(0.41) & 54.36(0.76) & 47.88(1.09) & 53.77 \\
 \bottomrule
\end{tabular}
\caption{F1-scores and their averages per test set from the argument component detection results of monolingual, monolingual post-processed (described as post), multilingual, multilingual post-processed (post), and cross-lingual experiments.} 
\label{table:arg_comp_res}
}
\end{center}
\end{table*}

\subsection{Few-shot Learning}

\textit{Few-shot learning} exploits limited annotated examples to train models, striking a balance between data scarcity and task complexity.
\citet{ma-etal-2022-template} proposed a template-free method for few-shot prompting for Named Entity Recognition (NER) by tackling it as a Language Model (LM) task with an Entity-oriented LM (EntLM) objective to solve the NER task. This avoids generating a new template corpus for each example in the data. We use this method in our experiments as it represents current state-of-the-art, at the time of writing, for sequence labelling in few-shot settings. Their approach consists of first retrieving class-specific words called \textit{label words} from a pre-trained model, and predict those label words at the position of each entity. They propose several ways of computing these label words, and in this work, we used the method based on the frequency, namely, we select the words that are the most frequent for the given class. We generate 10 such label words for each class.


Following EntLM's methodology \cite{ma-etal-2022-template}, for every \textit{k} sample three randomly sampled training sets are created. Training is then performed on each of these datasets over four iterations, and subsequently, sequence-level F1-scores and standard deviations are calculated.

In addition to the monolingual experiments, we also carry out multilingual experiments by combining all the French, Spanish, Italian, and English data. More specifically, we merge one sampled training file from each language of the k-percent sampling method. Evaluation is then conducted separately for each language.

Finally, we also compare EntLM with fine-tuning mBERT on few-shot settings.
 
\section{Results}\label{sec:results}
Following the completion of the experiments outlined in Section \ref{sec:method}, this section reports the obtained results using mBERT\footnote{Results obtained by training mDeBERTa-v3-base are in Appendix \ref{sec:appendix_mono_multi_cross_mdeberta}.}.

\begin{figure*}[ht]
    \centering
    \includegraphics[width=15cm]{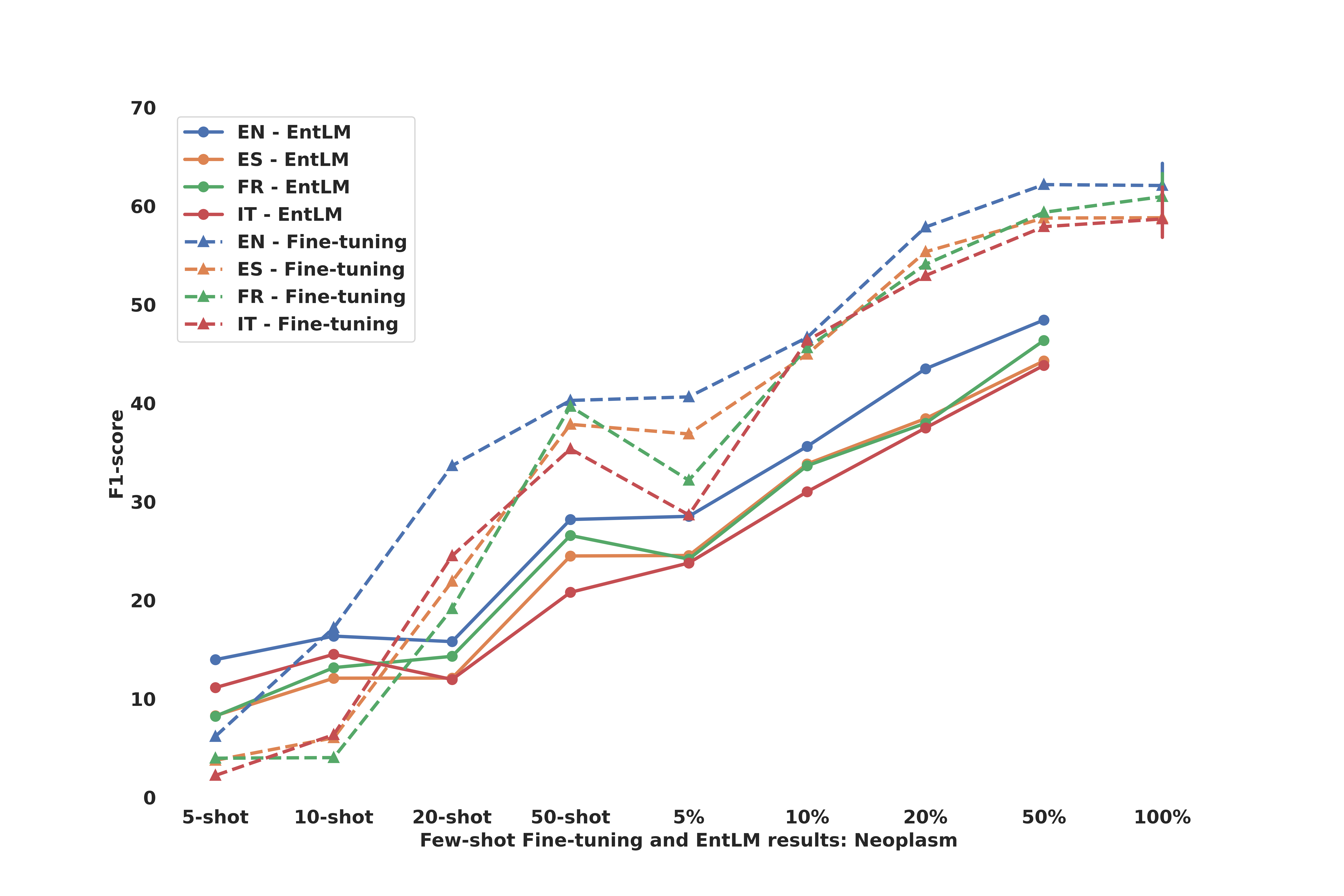}

    \caption{F1-score per \textit{k-shot} and \textit{k-percent} for Neoplasm from EntLM (dots and lines) and fine-tuning (triangles and dashed lines).}
    \label{fig:neo-entlm-ft}
\end{figure*}

\subsection{Model-transfer and Data-transfer}

Table \ref{table:arg_comp_res} displays the F1-scores derived from the argument component detection experiments using full in-domain data across all the experiments. The rows corresponding to the \textit{monolingual data-transfer} category present the results obtained from training and evaluating in the corresponding language. Similarly, \textit{multilingual data-transfer} refers to the merged training set consisting of all 4 languages and evaluating each language separately. \textit{Cross-lingual} refers to  model-transfer, namely, training in English and evaluating in the other 3 languages. The last column corresponds to the average between all the results per language across all test sets. For a fair comparison, the average of the cross-lingual model transfer includes the F1-score of the monolingual English.

Results show that, contrary to previous work on crosslingual transfer \cite{garcia-ferrero-etal-2022-model}, \textit{monolingual data-transfer} clearly outperforms \textit{cross-lingual model-transfer} for argument component detection. Another interesting point is that \textit{multilingual data-transfer} obtains the overall best results outperforming also the original English gold results. This means that \textit{data-transfer} may be employed as a cost-free data-augmentation technique. 

If we look at the results obtained when fine-tuning with the \textit{post-processed} data, results indicate that \textit{data-transfer} may be used in a fully automatic way, restricting the manual correction of the projected labels to the generation of evaluation sets.


\subsection{Few-shot}

Figure \ref{fig:neo-entlm-ft} reports the results of few-shot using both sampling methods (k-shot and k-percent) for the data trained by means of both EntLM and fine-tuning techniques.

The first point to mention is that \textit{data-transfer} also outperforms the few-shot prompting approach for sequence labelling proposed by EntLM. Furthermore, and quite surprisingly, fine-tuning remains competitive with respect to EntLM with the k-shot sampling while it is quite superior when tested on the percentage sampling. We hypothesized that k-percentage sampling produces better performance due to the higher proportion of \textit{outside} tokens. In fact, when fine-tuned with 20\% and 50\% of the data performance is comparable to that of data-transfer and model-transfer results.


 \begin{table}[ht]
    \begin{adjustbox}{max width=7.5cm}
    \centering
    \begin{tabular}{c|cccc}

         \textbf{EN} &   \textbf{Neoplasm}  &  \textbf{Glaucoma}   &   \textbf{Mixed} & \textbf{Avg.}      \\
         \hline
         5\%         & 41.92(8.39) &	47.60(9.43)	& 39.27(4.92) & 42.93 \\
         10\%        & 52.86(3.10) & 55.18(3.37) & 55.23(1.75) & 54.42 \\
         20\%        & 57.14(1.19) & 60.34(1.62) & 57.33(0.66) & 58.27 \\
        \hline
    
         \textbf{ES} &   \textbf{Neoplasm}  &  \textbf{Glaucoma}   &   \textbf{Mixed} & \textbf{Avg.}      \\
         \hline
         5\%         & 40.74(3.13) & 39.33(8.59) & 38.96(5.04) &  39.68 \\
         10\%        & 51.68(1.71) & 55.83(2.09) & 51.93(1.27) &  53.15\\
         20\%        & 59.04(0.58) & 57.39(1.92) & 55.27(1.78) &  57.23 \\
        \hline
         \textbf{FR} &  \textbf{Neoplasm}  &  \textbf{Glaucoma}   &   \textbf{Mixed} & \textbf{Avg.}      \\
         \hline
         5\%         & 37.45(7.38) & 42.42(3.51) & 29.01(4.48) & 36.29  \\
         10\%        & 50.46(1.75) & 53.03(2.17) & 50.71(1.63) & 51.40 \\
         20\%        & 57.45(1.29) & 55.70(2.55) & 56.57(1.35) & 56.57 \\
        \hline
         \textbf{IT} &  \textbf{Neoplasm}  &  \textbf{Glaucoma}   &   \textbf{Mixed} & \textbf{Avg.}     \\
         \hline
         5\%         & 37.07(8.43) & 47.96(2.95) & 37.59(9.49) & 40.87 \\
         10\%        & 50.78(1.61) & 53.91(3.65) & 49.48(4.81) & 51.39 \\
         20\%        & 55.85(1.54) & 57.53(2.92) & 54.75(1.61) & 56.04 \\
        \hline

    \end{tabular}

    \end{adjustbox} 
    \caption{F1-scores and standard deviation of multilingual few-shot fine-tuning mBERT with \textit{k-percent}.}
    \label{tab: fewshot_multi}
\end{table}


With respect to the multilingual experiments, one training sample from each k-percent sampling was merged into one training set, fine-tuned, and tested on each language (Table \ref{tab: fewshot_multi}). As observed in Figure \ref{fig:neo-entlm-ft}, fine-tuning with 50\% of the data (dash lines) produces results almost as high as 100\%. Furthermore, results demonstrate that merging 20\% of the data performs slightly worse than the model trained on the full data.

\section{Error Analysis}\label{sec:error}


In general, fine-tuning the model on the complete dataset often results in misclassifications with a tendency to assign \textit{Claim} labels in place of \textit{Premise}. Additionally, dealing with long sequences poses challenges in accurately identifying both boundaries and classes for the system. This pattern persists in zero-shot results, and it can be attributed to an inherent imbalance in the data, particularly in terms of the disparity between the number of \textit{Claim} and \textit{Premise} labels and the length of arguments in the sequences.

Each sequence predominantly corresponds to a single argument type, and instances where a sequence contains compound arguments, or when the argument span is only a proportion of the input, are less frequent. Consequently, in such examples, the most prevalent error involves misidentifying \textit{Claim} as \textit{Premise} and recognizing only one argument component in sequences with multiple components. These errors tend to occur more systematically in classifications under the zero-shot setting.

In \textit{k-shot} scenarios, the model consistently struggles to accurately identify both the correct spans and class labels. Furthermore, as the number of \textit{k} decreases, there is an increase in randomness in the assigned classes for each token, meaning that each token in a sequence may be classified differently. In particular, it is notable in the \textit{5}- and \textit{10}-shot. Under \textit{k-shot} the model struggles to predict \textit{B-} tokens. Whereas in the \textit{k-percent} the opposite occurs, namely, the model learns to predict the beginning of the sequence and fails to predict \textit{O} sequences correctly. Nevertheless, it is observed that as the amount of data increases, the quality of the predicted outcomes improves.

The described errors persist consistently in the case of EntLM. Additionally, when dealing with smaller training sets, the trained model tends to assign a single argument type to all examples in a document. As the value of \textit{k} increases, the randomness in predictions also grows proportionally. In other words, a larger amount of data leads to more unpredictable assigning of labels by the model on the token level.

A potential explanation for such behavior may be the selection of the label words. The concept involves computing label-specific words to later substitute them for few-shot learning. Given that the length of an argument is usually long enough, one selected label word may not represent the argument type correctly.



\section{Concluding Remarks}\label{sec:conclusion}
In this paper, we address the argument component identification task in the clinical domain in a scenario of lack of manually annotated data for languages other than English. We address the problem by applying cross-lingual transfer and prompt-based learning strategies in the AbstRCT corpus. Experimentation was facilitated by the generation of multilingual dataset by machine-translating and projecting the annotations of the original English AbstRCT into French, Italian, and Spanish.

The results of our experiments show that for long and structurally complex sequence labelling, as it is the case of component identification in Argument Mining, \textit{data-transfer} is a better strategy than model-transfer (RS1). Thus, fine-tuning mBERT in monolingual and multilingual settings showed results on an average of around 60 F1-scores for three test sets, outperforming any other approach, be that \textit{model-transfer} or \textit{few-shot} learning.

Furthermore, we have addressed the question of how much data is required to obtain similar results to those using the full data for training (RS2) by performing experiments in a few-shot learning approach. Thus, corpus splits of different granularity (5, 10, 20, and 50 shot or percentage) were used in the experimentation with EntLM and mBERT. The models in general perform better when trained with data sampled using the k-percent method (in comparison to k-shot) and by fine-tuning a pre-trained language model (instead of using a prompting method such as EntLM). Finally, empirical results indicate that by fine-tuning the multilingual model mBERT with 20\% of the data performance is competitive with data- and model-transfer approaches.

\section{Limitations}

Our evaluation focuses on Argument Mining, and it would be interesting to compare it with other sequence labelling tasks where the spans are also complex and heterogeneous. Furthermore, we experiment only in the medical domain, which may affect the results on the data-transfer method. We note, however, that our results clearly contradict previous results on model-transfer vs data-transfer previously obtained for other sequence labelling tasks \cite{garcia-ferrero-etal-2022-model}. Furthermore, we also demonstrate the importance of the data sampling method in few-shot scenarios \cite{ma-etal-2022-template}. In any case, it would be interesting to perform similar experiments on different domains and for other languages with the aim of providing a similar comparison to corroborate that our findings also apply more broadly.

\section*{Acknowledgements}

We would like to acknowledge the funding received by several MCIN/AEI/10.13039/501100011033 projects: {(i) Antidote (PCI2020-120717-2), and by European Union NextGenerationEU/PRTR; (ii) DeepKnowledge (PID2021-127777OB-C21) and ERDF A way of making Europe; (iii) LOTU (TED2021-130398B-C22) and European Union NextGenerationEU/PRTR; (iv) EDHIA (PID2022-136522OB-C22); (v) DeepMinor (CNS2023-144375) and European Union NextGenerationEU/PRTR. We also thank the European High Performance Computing Joint Undertaking (EuroHPC Joint Undertaking, EXT-2023E01-013) for the GPU hours. Anar Yeginbergen's PhD contract is part of the PRE2022-105620 grant, financed by MCIN/AEI/10.13039/501100011033 and by the FSE+.

\bibliography{anthology,custom}

\section*{Appendix}
\appendix

\section{EntLM and  Fine-tuning Results per Test Set }
\label{sec:appendix_fewshot}

Results for all test sets obtained from few-shot training using EntLM and fine-tuning mBERT are presented in Tables \ref{tab: fewshot_entlm_neoplasm} (Neoplasm), \ref{tab: fewshot_entlm_glaucoma} (Glaucoma), and \ref{tab: fewshot_entlm_mixed} (Mixed). 

\begin{table}[h]
    \begin{adjustbox}{max width=7.5cm}
    \centering
    \begin{tabular}{c|ccccc}
         \textbf{EntLM} & \textbf{EN} & \textbf{FR} & \textbf{IT} & \textbf{ES} & \textbf{Avg.}  \\

         \hline
         5 shot & 13.98(3.39) & 8.24(5.22) & 11.14(3.45) & 8.29(5.98) &  \textbf{11.16} \\
         10 shot & 16.37(4.73) & 13.17(3.41) & 14.52(3.91) & 12.10(3.43) & \textbf{14.04} \\
         20 shot & 15.81(3.69) & 14.32(3.23) & 11.97(2.46)  & 12.11(4.65) & 13.55 \\
         50 shot & 28.20(2.76) & 26.58(3.63) & 20.81(3.34)  & 24.50(2.49) & 25.02 \\
         \hline
        5\%	     & 28.52(2.93) & 24.18(2.17) & 23.79(2.19) & 24.55(2.05) & 25.26 \\
        10\%     & 35.62(3.04) & 33.66(1.49) & 31.02(1.71) & 33.84(2.09) & 33.54 \\
        20\%     & 43.49(2.25) & 37.97(2.02) & 37.49(1.73) & 38.44(1.67) & 39.35 \\
        50\%     & 48.44(2.06) & 46.37(2.55) & 43.84(1.83) & 44.29(1.92)& 45.74 \\
        \hline
        \textbf{mBERT} &  \textbf{EN} & \textbf{FR} & \textbf{IT} & \textbf{ES} & \textbf{Avg.}  \\
         \hline
         5 shot  & 6.19(3.28)  & 3.98(3.21) & 2.23(1.76) & 3.77(2.38) & 4.04 \\
         10 shot & 17.21(7.68) & 4.04(4.52) & 6.36(4.37)   & 6.05(6.19) & 8.42 \\
         20 shot & 33.66(9.94) & 19.15(7.48) & 24.51(8.99) & 21.92(9.91) & \textbf{24.81} \\
         50 shot & 40.28(3.24) & 39.67(4.88) & 35.36(5.77) & 37.85(5.57) & \textbf{39.29} \\
         \hline
        5\%	     & 40.64(7.32)  & 32.18(7.06) & 28.66(5.02) & 36.88(6.32) & \textbf{34.59} \\
        10\%     & 46.67(6.94) & 45.62(3.33) & 46.38(4.12) & 44.97(4.93) & \textbf{45.91} \\
        20\%     & 57.87(1.34) & 54.09(1.86) & 52.95(2.33) & 55.36(2.83) & \textbf{55.07} \\
        50\%     & 62.18(1.35) & 59.37(1.89) & 57.91(1.79) & 58.79(1.52) & \textbf{59.56} \\
    \end{tabular}
    \end{adjustbox} 
    \caption{Average F1-scores and standard deviation of few-shot  EntLM and fine-tuning on \textit{k-shot} and \textit{k-percent.} (Neoplasm)} 
    \label{tab: fewshot_entlm_neoplasm}
\end{table}


\begin{table}[h]
    \begin{adjustbox}{max width=7.5cm}
    \centering
    \begin{tabular}{c|ccccc}

         \textbf{EntLM} & \textbf{EN} & \textbf{FR} & \textbf{IT} & \textbf{ES} & \textbf{Avg.} \\
         \hline
         5 shot  & 12.87(6.44) & 8.47(4.01) & 10.76(3.85) & 9.06(7.04) & \textbf{10.29} \\
         10 shot & 18.09(3.75) & 13.18(3.63) & 14.59(5.88) & 12.35(5.59) & \textbf{14.55} \\
         20 shot & 17.86(6.86) & 19.75(3.41) & 16.18(3.81) & 14.87(2.34) & 17.17 \\
         50 shot & 26.38(2.99) & 28.39(2.06)& 25.06(3.48) & 27.34(2.15) & 26.79 \\
         \hline
        5\%	     & 30.36(2.19) & 27.33(3.06) & 25.51(3.92) & 26.25(1.86) & 27.36 \\
        10\%     & 38.28(2.73) & 36.19(3.03) & 33.38(6.08) & 33.45(2.79) & 35.33 \\
        20\%     & 48.51(2.06) & 40.69(2.02) & 41.22(1.99) & 39.97(3.60) & 42.59 \\
        50\%     & 51.98(3.07) & 48.71(2.73) & 50.44(2.60) & 50.87(2.58) & 50.50 \\
        \hline
        \textbf{mBERT} & \textbf{EN} & \textbf{FR} & \textbf{IT} & \textbf{ES} & \textbf{Avg.}  \\
         \hline
         5 shot & 4.19(5.91) & 3.65(2.95) & 1.76(1.16) & 3.79(2.63) & 3.35 \\
         10 shot  & 14.83(9.45)   & 5.43(2.83)   & 6.56(5.07)  & 7.70(8.29) & 8.63 \\
         20 shot  & 31.11(9.64)  & 23.79(7.74)  & 27.52(5.31)  & 23.24(6.26) & \textbf{26.41} \\
         50 shot  & 38.01(7.75) & 39.42(8.77)  & 38.60(7.49)  & 39.99(5.65) & \textbf{39.01} \\
         \hline
        5\%	      & 42.14(6.57)  & 41.56(6.54)  & 34.37(9.17)  & 39.18(6.12) & \textbf{39.31} \\
        10\%      & 44.73(8.78)  & 46.66(3.66)  & 47.14(6.43) & 43.71(6.20)   & \textbf{45.56} \\
        20\%      & 58.29(1.73)  & 55.74(2.97)  & 53.79(3.56)  & 54.99(3.12)   & \textbf{55.70} \\
        50\%      & 61.89(3.16)  & 62.66(2.41)  & 61.03(2.32)  & 61.79(3.05) & \textbf{61.84} \\
    \end{tabular}
    \end{adjustbox} 
    \caption{Average F1-scores and standard deviation of few-shot  EntLM and fine-tuning on \textit{k-shot} and \textit{k-percent.} (Glaucoma)}
    \label{tab: fewshot_entlm_glaucoma}
\end{table}

\begin{table}[h]
    \begin{adjustbox}{max width=7.5cm}
    \centering
    \begin{tabular}{c|ccccc}

         \textbf{EntLM} &  \textbf{EN} & \textbf{FR} & \textbf{IT} & \textbf{ES} & \textbf{Avg.} \\
         \hline
         5 shot & 11.75(3.91) & 9.16(5.53) & 11.09(4.22) & 7.24(6.94) & \textbf{9.81} \\
         10 shot & 17.25(4.35) & 14.48(4.28) & 14.17(4.31) & 12.20(3.45) & \textbf{14.53} \\
         20 shot & 14.87(5.31) & 18.37(2.39)  & 13.19(3.84)  & 11.99(4.19)  & 14.61 \\
         50 shot & 26.06(1.49)  & 24.17(2.96) & 22.75(3.05)  & 23.31(2.53) & 24.07 \\
         \hline
        5\%	     & 26.82(3.15)  & 22.76(2.45) & 21.99(1.76)  & 25.35(1.74)  & 24.23 \\
        10\%     & 34.73(2.64)  & 32.05(2.51) & 30.96(2.62)  & 32.56(2.70)  & 32.58 \\
        20\%     & 42.90(2.65) & 36.82(1.92) & 37.47(1.79)  & 37.98(2.22)  & 38.79 \\
        50\%     & 46.47(2.03) & 43.13(2.18) & 42.37(2.29)  & 43.73(2.12)  & 43.93 \\
        \hline
        \textbf{mBERT} &  \textbf{EN} & \textbf{FR} & \textbf{IT} & \textbf{ES} & \textbf{Avg.}  \\
         \hline
        
         5 shot & 2.80(4.42) & 4.03(2.42) & 1.53(1.59 & 2.38(2.56) & 2.69 \\
         10 shot & 13.59(6.09)  & 3.77(4.78)  & 8.45(5.30) & 6.62(6.94) & 8.11 \\
         20 shot & 31.41(8.41) & 22.26(8.57) & 26.38(6.25) & 26.80(9.01) & \textbf{26.71} \\
         50 shot & 40.97(3.43) & 39.79(6.65) & 35.78(7.41) & 38.94(6.46)  & \textbf{38.87} \\
         \hline
        5\%	     & 39.82(6.91) & 38.51(9.42) & 32.31(6.45) & 38.38(5.48) & \textbf{37.26} \\
        10\%     & 47.91(7.75) & 44.49(7.23) & 44.01(4.07) & 39.38(6.62) & \textbf{43.95} \\
        20\%     & 57.01(3.47) & 52.32(2.26) & 51.92(2.64) & 53.98(2.46)  & \textbf{53.81} \\
        50\%     & 61.44(1.97) & 59.19(2.72) & 57.61(2.23) & 58.51(2.07) & \textbf{58.19} \\
    \end{tabular}
    \end{adjustbox} 
    \caption{Average F1-scores and standard deviation of few-shot  EntLM and fine-tuning on \textit{k-shot} and \textit{k-percent.} (Mixed)} 
    \label{tab: fewshot_entlm_mixed}
\end{table}

\section{Results from Training on Multilingual Post-processed data}
\label{sec:appendix_multi_post}

In Table \ref{tab: fewshot_multi_post_processed}, the results of training on multilingual post-processed data (without manual correction) are reported.


 \begin{table}[ht]
    \begin{adjustbox}{max width=7.5cm}
    \centering
    \begin{tabular}{c|cccc}
    
         \textbf{ES} &   \textbf{Neoplasm}  &  \textbf{Glaucoma}   &   \textbf{Mixed} & \textbf{Avg.}      \\
         \hline
         5\%         & 36.95(8.49) & 37.75(17.94) & 38.71(7.99) & 37.80 \\
         10\%        & 45.11(7.21) & 46.89(2.82)  & 38.79(4.05) & 43.59 \\
         20\%        & 54.21(1.37) & 57.83(0.95)  & 52.30(1.21) & 54.78 \\
        \hline
         \textbf{FR} &  \textbf{Neoplasm}  &  \textbf{Glaucoma}   &   \textbf{Mixed} & \textbf{Avg.}      \\
         \hline
         5\%         & 44.57(2.29) & 45.25(5.29) & 46.93(4.73) & 45.58 \\
         10\%        & 42.86(8.49) & 39.34(6.96) & 42.77(3.29) & 41.66 \\
         20\%        & 53.21(2.27) & 56.89(0.96) & 53.01(1.14) & 54.37 \\
        \hline
         \textbf{IT} &  \textbf{Neoplasm}  &  \textbf{Glaucoma}   &   \textbf{Mixed} & \textbf{Avg.}     \\
         \hline
         5\%         & 44.57(2.53) & 49.16(3.41) & 39.11(6.56)  & 44.28 \\
         10\%        & 44.09(2.65) & 47.37(3.58) & 46.12(2.09)  &  45.86 \\
         20\%        & 54.78(0.56) & 55.69(1.13) & 52.41(1.09)  & 54.29 \\
        \hline

    \end{tabular}

    \end{adjustbox} 
    \caption{Average F1-scores and standard deviation of multilingual few-shot fine-tuning mBERT with \textit{k-percent} with post-processed data.}
    \label{tab: fewshot_multi_post_processed}
\end{table}

\section{Cross-lingual Few-shot results}
\label{sec:appendix_cross_few}

Results obtained from zero-shot cross-lingual few-shot experiments using \textit{k}=20 (shot and percent) with EntLM and fine-tuning mBERT are reported in Table \ref{tab: fewshot_cross_lingual}.

 \begin{table}[ht]
    \begin{adjustbox}{max width=7.5cm}
    \centering
    \begin{tabular}{c|cccc}
         

         \textbf{EntLM}  & \textbf{FR} & \textbf{IT} & \textbf{ES} & \textbf{Avg.} \\
         \hline
         \multicolumn{2}{}{} & \textbf{Neoplasm}   \\
         \hline
         20 shot & 5.67(1.99) & 6.68(3.18) & 9.71(3.89)  &  7.35 \\
         
        20\%     & 28.99(3.04) & 28.76(2.10) & 35.05(1.48)  & 30.93 \\
        \hline
        
        \multicolumn{2}{}{} & \textbf{Glaucoma}   \\
         \hline
      
         20 shot & 8.90(2.65) & 9.21(4.39) & 11.59(2.93)  &  9.90  \\
        20\%     & 31.51(2.34) & 31.73(3.92) & 36.11(2.01) &  33.12 \\
      \hline
        \multicolumn{2}{}{} & \textbf{Mixed} \\
         \hline
     
         20 shot & 8.11(2.79) & 6.90(2.74) & 11.37(2.85)  &  8.79 \\
        20\%     & 27.25(2.49) & 26.98(3.64) & 30.21(2.58)  & 28.15 \\
        \hline
        \textbf{mBERT}  & \textbf{FR} & \textbf{IT} & \textbf{ES} & \textbf{Avg.}  \\
         \hline
        
         \multicolumn{2}{}{} & \textbf{Neoplasm}   \\
         \hline
         20 shot & 10.07(7.62) & 20.42(8.27) & 17.92(8.68) & 16.04 \\
        20\%     & 46.69(0.29) & 47.86(5.75) & 51.79(3.81) &  48.78 \\
        \hline
        
        \multicolumn{2}{}{} & \textbf{Glaucoma}   \\
         \hline
         20 shot & 14.35(10.03) & 10.39(4.11) & 17.31(8.56) & 14.02   \\
        20\%     &  49.38(0.43) & 46.66(2.06) & 52.98(1.83) &  49.67  \\
        \hline
        \multicolumn{2}{}{} & \textbf{Mixed}  \\
         \hline
         20 shot & 10.17(6.42)  &  9.87(9.48) & 24.32(3.87) &  14.79  \\
        20\%     & 46.47(1.92)  & 47.94(0.66) & 49.86(2.64) &  48.09 \\

    \end{tabular}

    \end{adjustbox} 
    \caption{Average F1-scores and standard deviation of cross-lingual few-shot results using EntLM and fine-tuning mBERT with 20-shot and 20\%.}
    \label{tab: fewshot_cross_lingual}
\end{table}

\section{Monolingual, multilingual and cross-lingual mDeBERTa results}
\label{sec:appendix_mono_multi_cross_mdeberta}

Monolingual, multilingual, and cross-lingual mDeBERTa results.

\begin{table*}
\begin{center}
\footnotesize{
\begin{tabular}{cccccc} 
  \textbf{Test set} & \textbf{English} & \textbf{Spanish} & \textbf{French} & \textbf{Italian} & \textbf{Avg.} \\
  \toprule

 & \textbf{gold} & \multicolumn{3}{c}{\textbf{monolingual data-transfer}}\\
 \midrule
 
 Neoplasm & 59.29(0.57) & 58.46(2.53) & 60.66(1.99) & 58.19(1.11) & 59.15 \\
 Glaucoma & 64.38(1.21) & 64.84(0.69) & 63.17(1.45) & 67.39(1.04) & 64.95 \\
 Mixed    & 59.75(2.33) & 57.14(1.24) & 57.05(1.47) & 56.71(0.70) & 57.66 \\
 
\midrule

 & \textbf{gold} & \multicolumn{4}{c}{\textbf{monolingual data-transfer (post)}}\\
 \midrule
 
 Neoplasm & 59.29(0.57) & 58.83(1.44) & 55.39(1.20) & 58.19(1.26) & 57.93 \\
 Glaucoma & 64.38(1.21) & 63.12(2.15) & 60.36(0.65) & 64.38(2.56) & 63.06 \\
 Mixed    & 59.75(2.33) & 57.78(1.77) & 53.51(0.98) & 55.30(2.32) & 56.59 \\

\midrule 
\multicolumn{6}{c}{\textbf{multilingual data-transfer}} \\
\midrule
 Neoplasm & 63.16(0.66) & 61.21(0.47) & 56.44(1.69) & 54.16(1.62) & 58.74 \\
 Glaucoma & 69.53(1.24) & 67.92(1.17) & 64.62(0.58) & 60.58(1.33) & 65.66 \\
 Mixed    & 61.96(2.27) & 61.81(0.53) & 52.61(0.63) & 53.36(0.38) & 57.44 \\

 \midrule
 \multicolumn{6}{c}{\textbf{multilingual data-transfer (post)}} \\
 \midrule
  Neoplasm & 65.68(0.24) & 62.52(0.51) & 57.81(0.78) & 55.03(0.41) & 60.26 \\
  Glaucoma & 70.26(1.21) & 68.25(0.37) & 63.67(0.98) & 64.97(1.43) & 66.79 \\
  Mixed    & 65.66(0.88) & 60.76(1.18) & 57.88(0.62) & 57.31(3.30) & 60.40 \\
  \midrule
 \multicolumn{6}{c}{\textbf{cross-lingual model-transfer}} \\
 \midrule
 Neoplasm & - & 57.29(2.11) & 53.91(0.64) & 53.72(0.77) & 56.05 \\
 Glaucoma & - & 62.07(0.52) & 55.27(1.61) & 57.54(3.31) & 59.82 \\
 Mixed    & - & 54.95(2.03) & 50.63(0.30) & 52.35(1.57) & 54.42 \\
 \bottomrule
\end{tabular}
\caption{F1-scores and their averages per test set from the argument component detection results of monolingual, monolingual post-processed, multilingual, multilingual post-processed, and cross-lingual experiments using mDeBERTa.} 
\label{table:arg_comp_res_mdeberta}
}
\end{center}
\end{table*}

\end{document}